\pdfoutput=1

\documentclass[11pt]{article}

\usepackage[final]{acl}

\usepackage{times}
\usepackage{latexsym}

\usepackage[T1]{fontenc}

\usepackage[utf8]{inputenc}

\usepackage{microtype}

\usepackage{inconsolata}

\usepackage{graphicx}
\usepackage{colortbl}
\usepackage{subfiles}
\usepackage{booktabs}
\definecolor{darkgreen}{rgb}{0.0, 0.5, 0.0}
\definecolor{darkred}{rgb}{0.5, 0.0, 0.0}
\usepackage{enumitem}
\usepackage{amsfonts}

\title{SelfAug: Mitigating Catastrophic Forgetting in Retrieval-Augmented Generation via Distribution Self-Alignment}

\author{Yuqing Huang\textsuperscript{1}, Rongyang Zhang\textsuperscript{1}, Qimeng Wang\textsuperscript{2}, Chengqiang Lu\textsuperscript{2}, Yan Gao\textsuperscript{2}, \\
\textbf{Yi Wu\textsuperscript{2}, Yao Hu\textsuperscript{2}, Xuyang Zhi\textsuperscript{1}, Guiquan Liu\textsuperscript{1}\footnotemark[1], Xin Li\textsuperscript{1}, Hao Wang\textsuperscript{1}\thanks{Corresponding author}, Enhong Chen\textsuperscript{1}\footnotemark[1]} \\
  \textsuperscript{1}University of Science and Technology of China \quad
  \textsuperscript{2}Xiaohongshu Inc. \\
  \texttt{\{huangyuq,zhangry13,zxy\_zds\}@mail.ustc.edu.cn} \\
  \texttt{\{qimengwang,lusuo,wanjianyi,luyun2,xiahou\}@xiaohongshu.com} \\
  \texttt{\{gqliu,leexin,wanghao3,cheneh\}@ustc.edu.cn}} 

\begin{document}
\maketitle
\begin{abstract}
Recent advancements in large language models (LLMs) have revolutionized natural language processing through their remarkable capabilities in understanding and executing diverse tasks. While supervised fine-tuning, particularly in Retrieval-Augmented Generation (RAG) scenarios, effectively enhances task-specific performance, it often leads to catastrophic forgetting, where models lose their previously acquired knowledge and general capabilities. Existing solutions either require access to general instruction data or face limitations in preserving the model's original distribution. To overcome these limitations, we propose SelfAug, a self-distribution alignment method that aligns input sequence logits to preserve the model’s semantic distribution, thereby mitigating catastrophic forgetting and improving downstream performance. Extensive experiments demonstrate that SelfAug achieves a superior balance between downstream learning and general capability retention.
Our comprehensive empirical analysis reveals a direct correlation between distribution shifts and the severity of catastrophic forgetting in RAG scenarios, highlighting how the absence of RAG capabilities in general instruction tuning leads to significant distribution shifts during fine-tuning. 
Our findings not only advance the understanding of catastrophic forgetting in RAG contexts but also provide a practical solution applicable across diverse fine-tuning scenarios. Our code is publicly available at \textcolor{blue}{\url{https://github.com/USTC-StarTeam/SelfAug}}.
\end{abstract}

\section{Introduction}
Large language models (LLMs) like GPT \cite{achiam2023gpt}, PaLM \cite{chowdhery2023palm}, GLM \cite{glm2024chatglm}, and LLaMA \cite{touvron2023llama} have revolutionized NLP by learning complex linguistic patterns from extensive pre-training data, demonstrating excellence in contextual understanding and few-shot learning capabilities.

Supervised fine-tuning \cite{ouyang2022training, chung2024scaling} with general instruction datasets \cite{taori2023stanford, wang2022self} improves models' instruction following abilities but often inadequately addresses specialized domain tasks. Task-specific fine-tuning provides targeted solutions for specialized applications \cite{roziere2023code, yang2024qwen2, hui2024qwen2, luo2023wizardmath, jin2024genegpt, yin2024entropy, yin2024dataset, huang2024chemeval, shen2024optimizing, zhang2025td3}. Particularly, Retrieval-Augmented Generation (RAG) \cite{guu2020retrieval, lewis2020retrieval, gao2023retrieval, cai2022recent, chen2024benchmarking, shen2024exploring, shen2025genki, wu2024knowledge, gu-etal-2025-rapid, yu2025thought} enhances LLMs by incorporating external knowledge through retrieval, reducing hallucinations. Recent work \cite{yang2024crag, liu2024rag, zhang2024enhancing} improves how models utilize relevant information and handle insufficient information.

However, fine-tuning for downstream tasks introduces catastrophic forgetting \cite{french1999catastrophic, kemker2018measuring, shi2024continual, wu2024continual, luo2023empirical}, where models lose previously acquired knowledge and instruction-following abilities when adapting to new tasks. This causes performance deterioration across diverse applications. For example, a model fine-tuned on document extraction may generate structurally incorrect code, despite improved document parsing abilities. Recent research attributes this problem to distribution shift when models adapt to specialized task distributions during fine-tuning \cite{saha2021gradient, yang2024self}.

To address capability degradation, recent studies \cite{chen2024continual, bai2024pretrain, jin2024demystifying, huang2024mitigating} suggest incorporating general instruction data during downstream fine-tuning to maintain LLM's general capabilities. However, these strategies are limited by the scarcity of publicly available instruction datasets. Researchers have therefore explored alternative approaches that retain the model's original distribution without accessing general data. Instruction synthesis methods like MAGPIE \cite{xu2024magpie} use the model to generate instruction-response pairs for data replay, though they depend heavily on generation quality. Parameter constraint methods such as Orthogonal Loss \cite{wang2023orthogonal} enforce orthogonality between parameters but compromise downstream task performance. Knowledge reconstruction approaches like SDFT \cite{yang2024self} approximate the original distribution by regenerating responses from fine-tuning data but struggle with format-specific tasks, particularly when structured outputs like JSON are required. While each approach offers certain benefits, they all have limitations. These limitations underscore the need for more efficient solutions that better balance capability preservation and task adaptation.

To address the limitations of previous methods, we propose SelfAug, a novel approach that improves downstream performance while preserving the model's original capabilities. SelfAug is flexible and applicable to various fine-tuning scenarios. The core idea is to align the logits of input sequences during fine-tuning, leveraging the rich information in these logits generated by large language models during sequential processing. These logits reflect both learned knowledge and decision boundaries, helping to prevent catastrophic forgetting and ensuring the model’s behavior remains consistent while enabling it to learn new tasks \cite{hsu2022closer, sun2024logit}. Our analysis reveals that catastrophic forgetting is most severe in RAG scenarios, especially when longer reference documents are used. We find a direct link between larger distribution shift and greater forgetting, as well as a correlation between longer contexts and more significant shift. SelfAug effectively mitigates these issues, achieving performance similar to LoRA while maintaining the model's original abilities, demonstrating the value of aligning logits distributions \cite{hsu2022closer, sun2024logit}.  The main contributions of this work are as follows:

\begin{itemize}[leftmargin=*,align=left]

\item We introduce SelfAug, a novel self-alignment method based on logits. SelfAug aligns input sequence logits to overcome limitations of current methods regarding data access and parameter constraints. It requires no extra data or validation and avoids downstream performance loss caused by strict parameter updates.

\item We provide an empirical analysis of catastrophic forgetting in RAG scenarios, showing that missing RAG capability in general instruction tuning causes significant distribution shifts. We also find a direct link between distribution shift and catastrophic forgetting severity.

\item Experiments on various benchmarks demonstrate that SelfAug achieves superior downstream performance compared to existing methods while preserving the model’s original distribution and reducing catastrophic forgetting.

\end{itemize}

\begin{figure*}[ht]
  \centering
  \includegraphics[width=\linewidth]{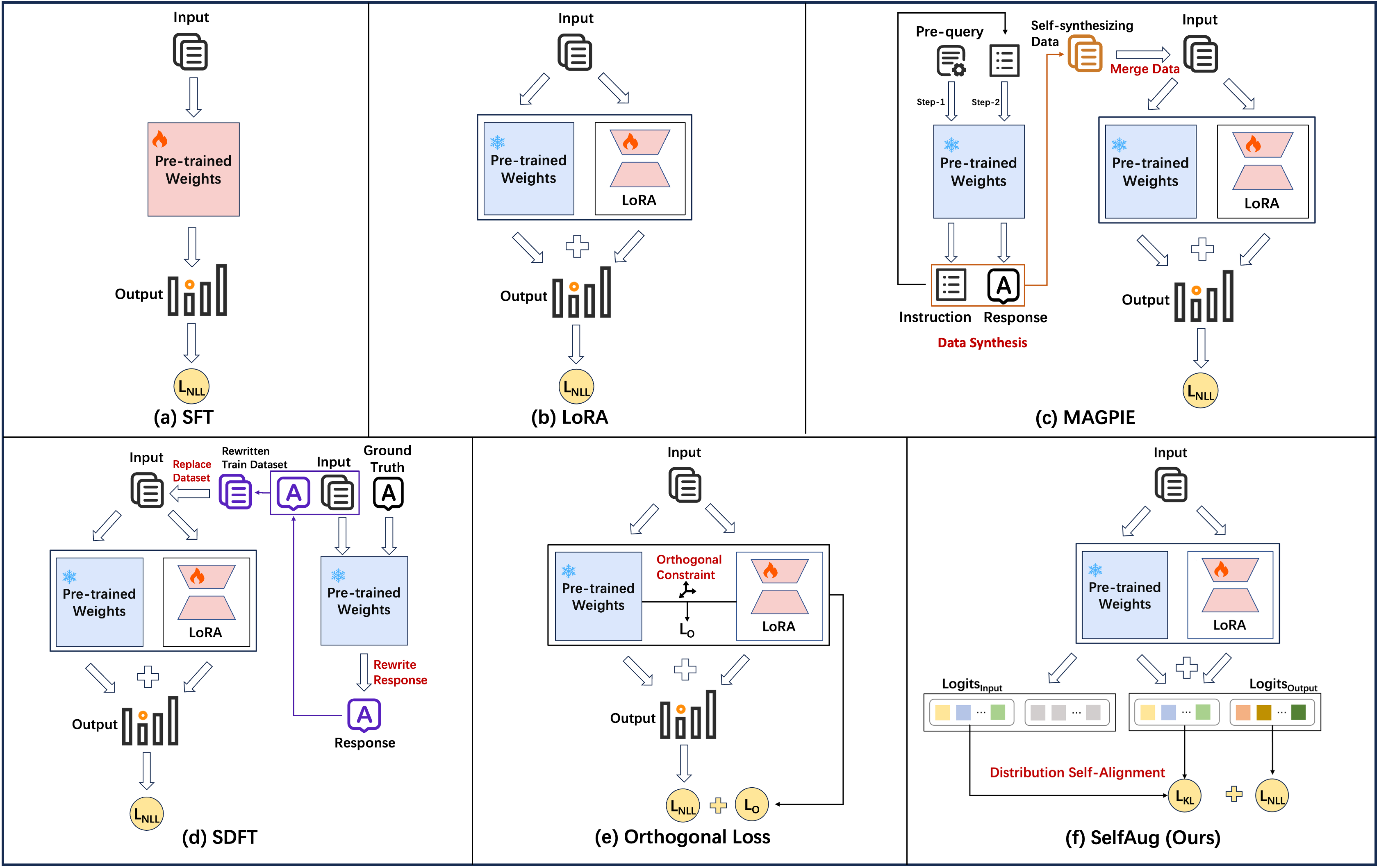}
  \caption{An illustration of full fine-tuning, LoRA, and methods for catastrophic forgetting mitigation. (a) SFT: Vanilla supervised fine-tuning with full parameter optimization. (b) LoRA: Parameter-efficient adaptation through low-rank decomposition. (c) MAGPIE: Self-synthesizing instruction-response pairs with pre-query templates for data replay. (d) SDFT: Fine-tuning with model-rewritten responses as optimized training dataset. (e) Orthogonal Loss: Imposing orthogonal constraints between LoRA modules and pre-trained parameters. (f) SelfAug: Self-distillation through input logits distribution alignment to preserve model's original capabilities.}
  \label{tab:method}
  \vskip -0.8em
\end{figure*}

\vskip -1em
\section{Related Works}
\vskip -0.2em
\subsection{Fine-Tuning}
Fine-tuning leverages the knowledge of pre-trained large models to improve their performance on specific downstream tasks. This approach has proven effective in areas such as mathematics \cite{luo2023wizardmath, yang2024qwen2, tang2024mathscale}, code \cite{roziere2023code, hui2024qwen2}, finance \cite{li2023large, wu2023bloomberggpt}, and healthcare \cite{yu2024enhancing}. Standard fine-tuning works by aligning the model’s output distribution with the downstream data through log-likelihood maximization. Although open-source LLMs are available for fine-tuning, training all parameters remains computationally expensive. Parameter-Efficient Fine-Tuning (PEFT) \cite{mangrulkar2022peft, han2024parameter} addresses this by optimizing fewer parameters. Low-Rank Adaptation (LoRA) \cite{hu2021lora} is a popular PEFT method that allows fine-tuning with significantly fewer trainable parameters. Recent research \cite{wang2023orthogonal, liu2024more, qiaolearn, kowsher2024propulsion} has focused on improving LoRA to increase performance with minimal training costs and to support multiple downstream tasks.

\vskip -0.2em
\subsection{Catastrophic Forgetting}
Fine-tuning models causes catastrophic forgetting as the model shift toward downstream task distributions and away from pre-training distributions. Traditional methods seek to balance performance across different tasks through various approaches. Parameter-constraining methods use regularization \cite{ni2024pace, xinruiforgetting} or selective parameter updates \cite{lin2024mitigating, alexandrov2024mitigating, marczak2025magmax, jin2024what, aggarwal2024exploring, franke2024preserving, panda2024lottery, zhang2024balancing, yangcorda, yang2024parameter}, but these limit downstream task performance. Mixture of Experts inspired approaches \cite{li2024mixlora, zhao2024mosld, le2024mixture, li2024moe} maintain general capabilities by using different parameters for different tasks but alter model structure and prevent parameter merging. Data replay techniques \cite{bai2024pretrain, jin2024will, aggarwal2024exploring, huang2024mitigating} preserve foundational knowledge but are constrained by the unavailability of pre-training data.

Among these, some methods focus on continual learning scenarios, emphasizing the balance of performance across multiple downstream tasks. These approaches typically employ mechanisms for knowledge retention or parameter constraints between tasks to minimize the interference of new task training on previously learned tasks. However, their primary goal is to optimize overall task performance, with limited attention to preserving the general capabilities of pre-trained models.
In contrast, our approach places greater emphasis on alleviating the forgetting of general capabilities in pre-trained models. We focus on maintaining the model’s inherent language understanding, reasoning, and knowledge abilities during fine-tuning, while simultaneously adapting to new task requirements. To address the limitations of the aforementioned methods, we propose a universal strategy aimed at systematically mitigating catastrophic forgetting in large language models during fine-tuning, allowing the model to retain its original capabilities while efficiently adapting to new tasks.

\vskip -0.2em
\subsection{Knowledge Distillation}
Knowledge distillation is widely used for model compression and performance improvement by transferring knowledge from a teacher model to a smaller student model. Early work \cite{hinton2015distilling, xie2018improving, liu2019structured, wang2020intra} focused on distilling knowledge from large models into smaller ones. Later studies applied knowledge distillation to various tasks \cite{shu2021channel, zhang2020improve, wang2019distilling}. For LLMs, the most common method \cite{mai2024fine, xu2024llavadi} uses KL divergence to reduce the difference between the teacher and student output distributions. Other methods \cite{hou2020dynabert, liang2023less} align their intermediate hidden states. Some approaches \cite{wang2022self, ding2023enhancing} transfer knowledge from closed-source API models by augmenting the training data.

Most existing knowledge distillation methods focus on transferring output sequences distributions to improve the downstream task performance of smaller models. In contrast, our method aims to reduce catastrophic forgetting during model fine-tuning by using the distribution of input sequences.

\vskip -0.2em
\section{Method}
In this section, we first outline the output logits of LLMs and the fine-tuning process. Subsequently, we introduce our SelfAug method and provide details on its implementation.

\subsection{Logits as Model Distribution Representations}
In LLM inference, input text undergoes several transformations to generate logits. Text is first tokenized into a sequence $x = [x_1, x_2, ..., x_n]$ and embedded into high-dimensional representations, then processed through multiple transformer layers to capture contextual relationships.

Finally, the model output is transformed into logits through a linear projection:
\vskip -0.5em
$$h_i=z_i^LW^T+b.$$
\vskip -0.2em
where $z_i^L \in \mathbb{R}^d$ represents the final layer hidden representation of the i-th token, $W^T \in \mathbb{R}^{d \times |V|}$ is the transpose of the projection matrix, and $b \in \mathbb{R}^{|V|}$ is the bias term. Each element in $h_i \in \mathbb{R}^{|V|}$ generates a corresponding score for each word in the vocabulary, reflecting the likelihood of selecting that word in the current context.

These logits are then converted to probability distributions via softmax for next-token prediction. The logit distribution encapsulates the linguistic patterns and semantic relationships learned during training \cite{jin2024what, lv2025costeer}.

\subsection{Fine-tuning: Aligning Model Distribution with Task Distribution}
While powerful, LLMs still require optimization for specific tasks. Fine-tuning is a crucial step that adjusts the model distribution to match the task data distribution. We denote the model to be fine-tuned as $M$ with parameters $\theta$, mapping instruction $x$ to output $y$.

Fine-tuning uses task-specific dataset $(x_t,y_t)\in D$ to update model parameters, aiming to minimize the negative log-likelihood loss:
\vskip -1.7em
$$\mathcal{L}_{NLL}(\theta) = -\sum_{(x_t, y_t) \in D}\log P(y_t \mid x_t; \theta).$$
\vskip -0.3em
By optimizing this function, the model's output distribution becomes closer to the true data distribution, with predicted outputs $\hat{y}_t$ more aligned with labels $y_t$. This process increases logits for target words and decreases them for others, making the model more suitable for specific task requirements.

\begin{table*}[htbp]
\centering
\caption{Results of Fine-tuning on Downstream Tasks in the RAG Domain (First CRAG, then RAG-Instruct). The CRAG benchmark employs a LLM-based ternary scoring mechanism (1: accurate, 0: missing, -1: incorrect) with overall performance represented by the mean score ranging from -1 to 1.}
\vskip -0.2em
\resizebox{\textwidth}{!} {
\begin{tabular}{l|ll|ll|lllll}
\toprule
\textbf{Dataset} & \textbf{Benchmark}& \textbf{Metric} & \textbf{Base} & \textbf{SFT} & \textbf{LoRA} & & & & \\
 & & & & & & \textbf{+MAGPIE} & \textbf{+SDFT} & \textbf{+Orthgonal} & \textbf{+SelfAug} \\
\midrule

& CRAG & score (\%) 
    & -13.11 & 9.59 & 8.76 
    & \underline{6.22} \textcolor{darkred}{\scriptsize 2.54$\downarrow$}
    & 4.34 \textcolor{darkred}{\scriptsize 4.42$\downarrow$}
    & 2.40 \textcolor{darkred}{\scriptsize 6.36$\downarrow$}
    & \textbf{10.94} \textcolor{darkgreen}{\scriptsize 2.18$\uparrow$} \\

& ChatRAGBench & F1 (\%) 
    & 24.04 & 25.92 & 31.90 
    & 33.56 \textcolor{darkgreen}{\scriptsize 1.66$\uparrow$}
    & 31.22 \textcolor{darkred}{\scriptsize 0.68$\downarrow$}
    & \underline{33.77} \textcolor{darkgreen}{\scriptsize 1.87$\uparrow$}
    & \textbf{34.46} \textcolor{darkgreen}{\scriptsize 2.56$\uparrow$} \\

\cmidrule{2-10} 

& BioASQ & F1 (\%) 
    & 66.76 & 59.41 & 59.70 
    & 62.06 \textcolor{darkgreen}{\scriptsize 2.36$\uparrow$}
    & \underline{64.71} \textcolor{darkgreen}{\scriptsize 5.01$\uparrow$}
    & 62.35 \textcolor{darkgreen}{\scriptsize 2.65$\uparrow$}
    & \textbf{65.00} \textcolor{darkgreen}{\scriptsize 5.30$\uparrow$} \\

& OmniEval & F1 (\%) 
    & 66.05 & 42.58 & 51.64 
    & \underline{54.71} \textcolor{darkgreen}{\scriptsize 3.07$\uparrow$}
    & 48.87 \textcolor{darkred}{\scriptsize 2.77$\downarrow$}
    & 49.53 \textcolor{darkred}{\scriptsize 2.11$\downarrow$}
    & \textbf{57.30} \textcolor{darkgreen}{\scriptsize 5.66$\uparrow$} \\

\cmidrule{2-10} 

& MATH & accuracy (\%) 
    & 69.56 & 53.84 & 65.64 
    & 68.36 \textcolor{darkgreen}{\scriptsize 2.72$\uparrow$}
    & \underline{69.26} \textcolor{darkgreen}{\scriptsize 3.62$\uparrow$}
    & 68.78 \textcolor{darkgreen}{\scriptsize 3.14$\uparrow$}
    & \textbf{69.46} \textcolor{darkgreen}{\scriptsize 3.82$\uparrow$} \\

\textbf{CRAG} & HumanEval & pass@1 (\%) 
    & 79.88 & 76.83 & 78.05 
    & 78.05 \textcolor{darkgreen}{\scriptsize 0.00$\uparrow$}
    & 76.83 \textcolor{darkred}{\scriptsize 1.22$\downarrow$}
    & \textbf{79.88} \textcolor{darkgreen}{\scriptsize 1.83$\uparrow$}
    & \underline{79.27} \textcolor{darkgreen}{\scriptsize 1.22$\uparrow$} \\

& IFEval & accuracy (\%) 
    & 71.90 & 45.10 & 48.80 
    & 58.04 \textcolor{darkgreen}{\scriptsize 9.24$\uparrow$}
    & 54.71 \textcolor{darkgreen}{\scriptsize 5.91$\uparrow$}
    & \textbf{63.77} \textcolor{darkgreen}{\scriptsize 14.97$\uparrow$}
    & \underline{62.11} \textcolor{darkgreen}{\scriptsize 13.31$\uparrow$} \\

\cmidrule{2-10} 

& MMLU & accuracy (\%) 
    & 74.23 & 72.24 & 73.72 
    & 73.56 \textcolor{darkred}{\scriptsize 0.16$\downarrow$}
    & 73.29 \textcolor{darkred}{\scriptsize 0.43$\downarrow$}
    & \textbf{74.45} \textcolor{darkgreen}{\scriptsize 0.73$\uparrow$}
    & \underline{74.04} \textcolor{darkgreen}{\scriptsize 0.32$\uparrow$} \\

& ARC-C & accuracy (\%) 
    & 86.78 & 85.08 & 88.47 
    & 88.47 \textcolor{darkgreen}{\scriptsize 0.00$\uparrow$}
    & \underline{89.83} \textcolor{darkgreen}{\scriptsize 1.36$\uparrow$}
    & 89.15 \textcolor{darkgreen}{\scriptsize 0.68$\uparrow$}
    & \textbf{90.17} \textcolor{darkgreen}{\scriptsize 1.70$\uparrow$} \\

& HellaSwag & accuracy (\%) 
    & 85.48 & 83.72 & 84.55 
    & 83.68 \textcolor{darkred}{\scriptsize 0.87$\downarrow$}
    & 82.54 \textcolor{darkred}{\scriptsize 2.01$\downarrow$}
    & \textbf{85.11} \textcolor{darkgreen}{\scriptsize 0.56$\uparrow$}
    & \underline{83.73} \textcolor{darkred}{\scriptsize 0.82$\downarrow$} \\

\cmidrule{2-10} 

\rowcolor{gray!20}
& Average & & 71.57 & 63.73 & 67.22 
    & 68.89 \textcolor{darkgreen}{\scriptsize 1.67$\uparrow$}
    & 68.02 \textcolor{darkgreen}{\scriptsize 0.80$\uparrow$}
    & \underline{69.36} \textcolor{darkgreen}{\scriptsize 2.14$\uparrow$}
    & \textbf{70.73} \textcolor{darkgreen}{\scriptsize 3.51$\uparrow$} \\

\midrule

& CRAG & score (\%) &
    -13.11 & -13.63 & -7.19 
    & \underline{-11.16} \textcolor{darkred}{\scriptsize 3.97$\downarrow$}
    & -17.00 \textcolor{darkred}{\scriptsize 9.81$\downarrow$}
    & -11.99 \textcolor{darkred}{\scriptsize 4.80$\downarrow$}
    & \textbf{-6.22} \textcolor{darkgreen}{\scriptsize 0.97$\uparrow$} \\

& ChatRAGBench & F1 (\%) &
    24.04 & 34.92 & 34.82 
    & \underline{33.59} \textcolor{darkred}{\scriptsize 1.23$\downarrow$}
    & 29.90 \textcolor{darkred}{\scriptsize 4.92$\downarrow$}
    & 29.16 \textcolor{darkred}{\scriptsize 5.66$\downarrow$}
    & \textbf{35.44} \textcolor{darkgreen}{\scriptsize 0.62$\uparrow$} \\

\cmidrule{2-10} 

& BioASQ & F1 (\%) &
    66.76 & 68.82 & 66.47 
    & \underline{66.76} \textcolor{darkgreen}{\scriptsize 0.29$\uparrow$}
    & 66.18 \textcolor{darkred}{\scriptsize 0.29$\downarrow$}
    & 64.41 \textcolor{darkred}{\scriptsize 2.06$\downarrow$}
    & \textbf{70.00} \textcolor{darkgreen}{\scriptsize 3.53$\uparrow$} \\

& OmniEval & F1 (\%) &
    66.05 & 66.37 & 66.62 
    & \textbf{67.68} \textcolor{darkgreen}{\scriptsize 1.06$\uparrow$}
    & 64.98 \textcolor{darkred}{\scriptsize 1.64$\downarrow$}
    & 66.84 \textcolor{darkgreen}{\scriptsize 0.22$\uparrow$}
    & \underline{67.58} \textcolor{darkgreen}{\scriptsize 0.96$\uparrow$} \\

\cmidrule{2-10} 

\textbf{RAG-} & MATH & accuracy (\%) &
    69.56 & 69.64 & 69.88 
    & 68.12 \textcolor{darkred}{\scriptsize 1.76$\downarrow$}
    & 69.82 \textcolor{darkred}{\scriptsize 0.06$\downarrow$}
    & \textbf{70.74} \textcolor{darkgreen}{\scriptsize 0.86$\uparrow$}
    & \underline{70.02} \textcolor{darkgreen}{\scriptsize 0.14$\uparrow$} \\

\textbf{Instruct} & HumanEval &pass@1 (\%)& 
    79.88 & 46.34 & 76.83 
    & \textbf{79.88} \textcolor{darkgreen}{\scriptsize 3.05$\uparrow$}
    & 76.22 \textcolor{darkred}{\scriptsize 0.61$\downarrow$}
    & \underline{79.27} \textcolor{darkgreen}{\scriptsize 2.44$\uparrow$}
    & \underline{79.27} \textcolor{darkgreen}{\scriptsize 2.44$\uparrow$} \\

& IFEval & accuracy (\%) &
    71.90 & 55.64 & 63.77 
    & 64.32 \textcolor{darkgreen}{\scriptsize 0.55$\uparrow$}
    & 66.73 \textcolor{darkgreen}{\scriptsize 2.96$\uparrow$}
    & \textbf{73.20} \textcolor{darkgreen}{\scriptsize 9.43$\uparrow$}
    & \underline{68.02} \textcolor{darkgreen}{\scriptsize 4.25$\uparrow$} \\

\cmidrule{2-10} 

& MMLU & accuracy (\%) &
    74.23 & 73.61 & 73.36 
    & 72.96 \textcolor{darkred}{\scriptsize 0.40$\downarrow$}
    & 73.28 \textcolor{darkred}{\scriptsize 0.08$\downarrow$}
    & \textbf{74.61} \textcolor{darkgreen}{\scriptsize 1.25$\uparrow$}
    & \underline{73.66} \textcolor{darkgreen}{\scriptsize 0.30$\uparrow$} \\

& ARC-C & accuracy (\%) &
    86.78 & 90.85 & 90.17 
    & 86.78 \textcolor{darkred}{\scriptsize 3.39$\downarrow$}
    & \underline{89.49} \textcolor{darkred}{\scriptsize 0.68$\downarrow$}
    & 88.14 \textcolor{darkred}{\scriptsize 2.03$\downarrow$}
    & \textbf{92.20} \textcolor{darkgreen}{\scriptsize 2.03$\uparrow$} \\

& HellaSwag & accuracy (\%) &
    85.48 & 82.21 & 83.45 
    & 82.36 \textcolor{darkred}{\scriptsize 1.09$\downarrow$}
    & 82.98 \textcolor{darkred}{\scriptsize 0.47$\downarrow$}
    & \textbf{85.82} \textcolor{darkgreen}{\scriptsize 2.37$\uparrow$}
    & \underline{84.93} \textcolor{darkgreen}{\scriptsize 1.48$\uparrow$} \\

\cmidrule{2-10} 

\rowcolor{gray!20}
& Average & & 71.57 & 66.30 & 70.77 
    & 70.36 \textcolor{darkred}{\scriptsize 0.41$\downarrow$}
    & 70.13 \textcolor{darkred}{\scriptsize 0.64$\downarrow$}
    & \underline{71.89} \textcolor{darkgreen}{\scriptsize 1.12$\uparrow$}
    & \textbf{72.51} \textcolor{darkgreen}{\scriptsize 1.74$\uparrow$} \\

\bottomrule
\end{tabular}
}
\label{tab:main_result}
\vskip -1em
\end{table*}

\subsection{SelfAug: Preserving Model Distribution via Input Logits}
From a Bayesian perspective, model parameters $\theta$ exist within a probability distribution where pre-training establishes the prior distribution $p(\theta)$ that confers general abilities. During fine-tuning on a new dataset $D$, these parameters update to a posterior distribution $p(\theta \mid D)$ to adapt to the current task. However, when this update relies exclusively on the new dataset, the posterior may diverge substantially from the original prior, leading to catastrophic forgetting where the model loses its general knowledge and generalization ability. To mitigate this issue, we explicitly define the prior $p(\theta)$ as a distribution that remains close to the original model distribution, constraining it through the distributional distance between the fine-tuned model $f_\theta$ and the original model $f_{\theta_0}$, as follows:

\vskip -0.8em
$$p(\theta)=exp(-\alpha \cdot Dist(f_{\theta},f_{\theta_0}))$$
where $Dist(f_{\theta},f_{\theta_0})$ denotes the distance between the distributions from the fine-tuned model and the original model, and $\alpha$ is a hyperparameter that controls the strength of this constraint. Therefore, the objective for optimizing the parameter posterior distribution during fine-tuning is as follows:
\vskip -0.8em
$$\theta^*=\mathop{argmax}\limits_\theta\ p(\theta \mid D)$$
\vskip -1.7em
$$=\mathop{argmin}\limits_\theta -log\ 
 p(D \mid \theta)+\alpha \cdot Dist(f_{\theta},f_{\theta_0})$$
\vskip -2.1em
$$=\mathop{argmin}\limits_\theta\ \mathcal{L}_{NLL}+\alpha \cdot Dist(f_{\theta},f_{\theta_0})$$
\vskip -0.8em
This design ensures that while the model parameters adapt to new data, their distribution does not deviate too far from that of the original model, which helps improve the model’s adaptability to new tasks and effectively preserves the original knowledge and generalization ability.

We propose the SelfAug, which aims to enhance performance on downstream tasks while maintaining the model's original distribution, as shown in Figure \ref{tab:method}(g). We leverage the characteristic of LLMs in receiving sequential inputs, where the model produces logits for both input sequence $x_t$ and the response sequence $y_t$, which together represent the original output distribution. Our key insight is using the original model's input sequence logits as a reference during fine-tuning. We measure the distribution difference between the original model be $M_{o}$ and the fine-tuning model be $M_{ft}$ using Kullback-Leibler divergence. For any input $x_t$, with logits $h_{o}(x_t)$ and $h_{ft}(x_t)$from respective models, we define the KL loss as:
\vskip -1.6em
$$Dist(f_{\theta},f_{\theta_0})=\mathcal{L}_{KL}=D_{KL}(p_{ft}(x_t)\mid\mid p_{o}(x_t)).$$
\vskip -0.4em
where $p_{o}(x_t)=softmax(h_{o}(x_t))$ and $p_{ft}(x_t)$ $=softmax(h_{ft}(x_t))$. The total loss function combines the negative log-likelihood loss $\mathcal{L}_{NLL}$ for the response sequences and the KL divergence loss:
\vskip -0.6em
$$\mathcal{L}_{total}=\mathcal{L}_{NLL}+\alpha\mathcal{L}_{KL}.$$
\vskip -0.2em
where $\alpha$ is a hyperparameter that balances the importance of the two loss terms.

SelfAug aligns the distribution of the original model through the logits of input sequences during the fine-tuning process. For each training pair $(x_t,y_t)$, the model not only learns the data distribution of downstream tasks through the response sequence $y_t$, but also maintains the distribution of the original model through the logits of the input sequence $x_t$. This integration of dual distributions effectively alleviates the catastrophic forgetting problem. Compared to methods requiring replay of original data or generation of responses, SelfAug offers the advantage of not needing additional data or complex response validation steps, thereby simplifying the implementation process and reducing computational overhead.

\vskip -0.2em
\section{Experiment}

To evaluate the effectiveness of SelfAug and its impact across different scenarios, we aim to answer the following research questions:
\begin{itemize}[leftmargin=*,itemsep=0pt,topsep=5pt,parsep=2pt,align=left]
    \item \textbf{RQ1}: How does SelfAug perform compared with the state-of-the-art methods?
    \item \textbf{RQ2}: How does constrained distributional shift mitigate catastrophic forgetting?
    \item \textbf{RQ3}: How do different components influence SelfAug?
    \item \textbf{RQ4}: How does SelfAug perform across varying context lengths and model configurations?
\end{itemize}

\subsection{Experimental Setup}
\paragraph{\textbf{Baselines.}} In our empirical investigation, we conduct extensive experiments using Qwen2.5-7B-Instruct \cite{qwen2.5} as our base model for fine-tuning. To systematically evaluate the effectiveness of our proposed method, we compare it with representative approaches from four major categories: instruction synthesis methods, knowledge reconstruction approaches, model modifications, and parameter constraint methods. We consider the following five baseline methods as our comparative benchmarks, as shown in Figure \ref{tab:method}(a)-(e):

\begin{itemize}[leftmargin=*,align=left]
\item \textbf{Vanilla Fine-Tuning}: We provide experimental results for both full-parameter fine-tuning and Low-Rank Adaptation (LoRA) \cite{hu2021lora} fine-tuning for comparison.

\item \textbf{MAGPIE} \cite{xu2024magpie}: In this approach, the LLM autonomously generates instructions when provided with pre-query templates as input, and subsequently produces corresponding responses for these instructions. The synthesized instruction-response pairs are utilized as alternative training samples for general instruction fine-tuning during data replay.

\item \textbf{SDFT} \cite{yang2024self}: This method bridges the distribution gap by fine-tuning with a dataset generated from the model’s distribution. The guiding model regenerates responses and validates their correctness to ensure alignment with the original data distribution.

\item \textbf{Orthogonal Loss}: Inspired by the concept of O-LoRA \cite{wang2023orthogonal}, this approach constrains the parameters of the LoRA modules to be orthogonal to the original model parameters, with the goal of minimizing the impact of fine-tuning on the model's distribution.
\end{itemize}

\vskip -0.6em
\paragraph{\textbf{Datasets.}} Our experimental evaluation consists of three main components: RAG capability, downstream task, and foundation knowledge. Each component assesses the performance of our approach across distinct domains.

\begin{itemize}[leftmargin=*,align=left]
\item \textbf{RAG Ability Evaluation.} We focus on enhancing RAG capabilities: document-based information retrieval and question answering, robustness against irrelevant or noisy documents, and the ability to abstain from answering given erroneous queries or insufficient context. For validation, we fine-tune our models on two datasets: \textbf{CRAG} \cite{yang2024crag} and \textbf{RAG-Instruct} \cite{liu2024rag}, and evaluate on two benchmarks: \textbf{CRAG} and \textbf{ChatRAGBench} \cite{liu2024chatqa}. 

\item \textbf{Domain-specific RAG Evaluation.} We evaluate RAG capabilities in the biomedical and financial domains using \textbf{BioASQ} \cite{nentidis2024overview} and \textbf{OmniEval} \cite{wang2024omnieval}. 

\item \textbf{Foundational Ability Evaluation.} For mathematical reasoning, we utilize the \textbf{MATH} \cite{hendrycks2021measuring}, which comprises 12,500 mathematics problems. For code generation Ability, we employ the \textbf{HumanEval} \cite{chen2021evaluating} to evaluate the model's programming proficiency. We evaluate the model's instruction-following ability using \textbf{IFEval} \cite{zhou2023instruction}, which assesses the model's capability to follow various types of instructions.

\item \textbf{General Knowledge Evaluation.} To evaluate the preservation of foundation knowledge, we employ three established benchmarks: \textbf{MMLU} \cite{hendrycks2020measuring}, \textbf{ARC} \cite{clark2018think}, and \textbf{HellaSwag} \cite{zellers2019hellaswag}. 
\end{itemize}
The evaluations on the MATH, HumanEval, MMLU, ARC, and HellaSwag datasets are conducted using the standardized OpenCompass \cite{2023opencompass} evaluation framework to ensure consistency and reproducibility.

\paragraph{Implementation Details.} For the CRAG dataset, we strictly adhere to the official configuration, utilizing the validation set for fine-tuning and the public test set for evaluation under Task 1 settings.  
Unless otherwise specified, we set the KL divergence loss weight in SelfAug to 0.5 in experiments, as our ablation studies confirm that 0.5 is a reasonable value. To ensure fair comparisons across tasks and metrics, score normalization is applied when computing the overall average performance. We conducted five repeated experiments to obtain the best value and determined the above hyperparameters through a hyperparameter grid search. The experiment was conducted using 4 A100 GPUs. More details are provided in Appendix \ref{sec:appendix_setup}.

\vskip -0.2em
\subsection{Overall Performance Evaluation (RQ1)}
We first evaluated the effectiveness of our proposed SelfAug method, which can maintain the performance of LLMs on downstream task learning while mitigating catastrophic forgetting during the fine-tuning process. Specifically, we conducted fine-tuning on the RAG dataset to assess the impact on the model's performance in both RAG tasks and other general capability tasks. Additionally, we observed that fine-tuning downstream tasks significantly affected the model's instruction-following abilities, whereas the impact on the model's knowledge was relatively mild. The evaluation results are presented in Table \ref{tab:main_result}.

\vskip -0.2em
\subsubsection{\textbf{SelfAug Effectively Mitigated Catastrophic Forgetting.}}
Our experimental results show that while fine-tuning improves downstream task performance, it also induces distribution shift that impair other capabilities. After applying LoRA fine-tuning to the CRAG dataset, the IFEval accuracy dropped to 48.80, indicating significant catastrophic forgetting. Although MAGPIE and SDFT were effective in mitigating catastrophic forgetting, SelfAug demonstrated superior performance in this regard. Orthogonal Loss, while mitigating catastrophic forgetting through strict orthogonal constraints that limit parameter updates to preserve generalization, significantly compromised downstream task performance. In contrast, SelfAug aligns the model’s semantic distribution without directly restricting parameter updates, allowing greater flexibility. It effectively mitigates forgetting while achieving exceptional results in downstream task learning, outperforming LoRA on targeted tasks. Among all the methods evaluated, SelfAug strikes the optimal balance between downstream task learning and catastrophic forgetting mitigation, achieving the highest average performance across all evaluation metrics.

\vskip -0.2em
\subsubsection{\textbf{The Impact on the Model's Knowledge is Slight.}}
Table \ref{tab:main_result} illustrates the results of the foundation knowledge assessment after fine-tuning with downstream tasks. While fine-tuning substantially deteriorates the model's instruction-following ability, its foundation knowledge retention remains remarkably robust. The performance across various foundation knowledge benchmarks exhibits minimal degradation after fine-tuning, with certain methodologies even demonstrating enhanced performance. These findings suggest that catastrophic forgetting in LLMs predominantly manifests through the degradation of instruction-following abilities rather than the erosion of foundation knowledge. This observation is also supported by other studies \cite{zhang2024dissecting, yang2024self}.

\begin{figure}[t]
  \centering
  \includegraphics[width=\linewidth]{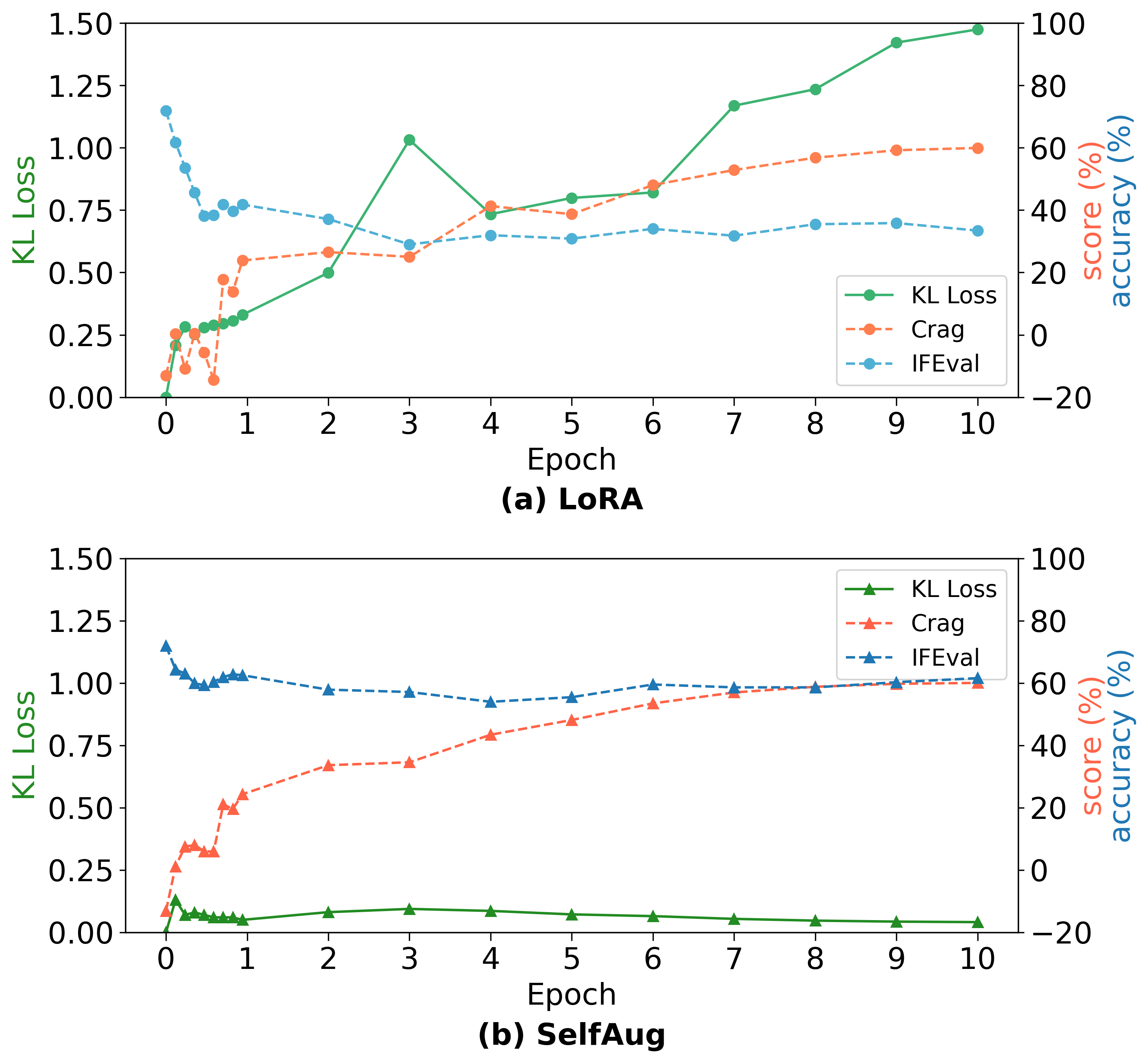}
  \caption{Epoch-wise Performance and Logits Divergence. KL Loss measures the distribution shift of model output logits, IFEval evaluates instruction-following ability catastrophic forgetting, and CRAG represents downstream task performance. LoRA exhibits increasing shift and forgetting, while SelfAug maintains stable performance through effective distribution constraints.}
  \label{tab:diff}
  \vskip -0.6em
\end{figure}

\vskip -0.4em
\subsection{Distribution Shift and Catastrophic Forgetting (RQ2)}

In this section, we explore how RAG task performance, instruction-following abilities, and distribution shift evolve over the course of training. After incorporating SelfAug, by imposing constraints on the distribution shift, we can alleviate catastrophic forgetting while preserving RAG task performance.

\vskip -0.4em
\subsubsection{\textbf{Distribution Shift Induced Catastrophic Forgetting.}}
We trained the LLM for 10 epochs and visualized its performance across the CRAG training set, IFEval datasets, as well as changes in KL Loss. As shown in Figure \ref{tab:diff}(a), increasing the number of training epochs progressively improves both the performance of model on Crag and logits distribution shift. At the same time, instruction-following ability suffers from a severe decline. This phenomenon reveals a strong correlation between the magnitude of distribution shift and the severity of catastrophic forgetting. The results demonstrate that continued training leads to increases in both RAG performance and logits distribution divergence, while degrading general capabilities. 

\vskip -0.4em
\subsubsection{\textbf{Effectiveness of SelfAug in Mitigating Distribution Shift.}}
Based on these observations, SelfAug leverages logits distribution self-alignment to constrain distribution shift during model training, effectively mitigating catastrophic forgetting. As demonstrated in Figure \ref{tab:diff}(b), after applying the SelfAug constraint, the KL divergence of model logits significantly decreases and remains at a stable level. Furthermore, the degradation of instruction-following ability is notably suppressed, confirming the effectiveness of our method in mitigating catastrophic forgetting phenomena. Notably, while mitigating catastrophic forgetting, SelfAug does not compromise the model's performance on training data, demonstrating a well-balanced trade-off between maintaining downstream task learning capabilities and preventing catastrophic forgetting.

\begin{table}[t]
\centering
\caption{Performance Comparison of Constraints Using Different Layer Outputs.}
\resizebox{0.5\textwidth}{!} {
\begin{tabular}{lc|lc}
\toprule
\textbf{Method}  & \textbf{IFEval} & \textbf{Method}  & \textbf{IFEval} \\
\midrule
LoRA & 48.80 & LoRA & 48.80 \\
\midrule
+ Attention Q & 47.13 & + Attention All & 50.46\\
+ Attention K & 50.09 & + FFN & 51.02\\
+ Attention V & 48.24 & + All layers & 49.35\\
+ Attention O & 47.50 & + SelfAug (Ours)  & \textbf{62.11} \\

\bottomrule
\end{tabular}
}
\vskip -0.6em
\label{tab:layers}
\end{table}

\subsection{Ablation Study (RQ3)}
Since distribution shift can occur on features at any module within the model, the effectiveness of SelfAug might be influenced by two factors: the location where constraints are applied and the strength of the constraints. Therefore, in the ablation study, we will focus primarily on these two aspects.

\subsubsection{\textbf{The Impact of Loss Position.}}
Previous studies have explored knowledge distillation through intermediate features, but our systematic comparison across different components of Transformer blocks reveals that distilling at the logits layer consistently achieves superior performance, as shown in Table \ref{tab:layers}. From the perspective of information bottleneck theory, as data propagates through the network, information is progressively filtered to emphasize task-relevant features, and the final logits primarily retain essential semantic content. Thus, distillation at this layer not only aligns the model more closely with task-relevant information but also improves generalization and robustness, whereas intermediate layers often mix relevant and irrelevant signals, introducing unnecessary complexity. Regarding alignment strategies, aligning output sequence logits may cause interference with downstream learning because both operate in the output space, while aligning intermediate features forces the model to replicate computations at every layer of the original model, overly constraining its representational capacity. In contrast, input sequence logits alignment provides the probability distribution over each token in the input, better capturing semantic representations while being disentangled from downstream objectives. This design avoids interference, enabling the model to adapt to new tasks while preserving its general capabilities and mitigating catastrophic forgetting.

\begin{figure}[t]
  \centering
  \includegraphics[width=\linewidth]{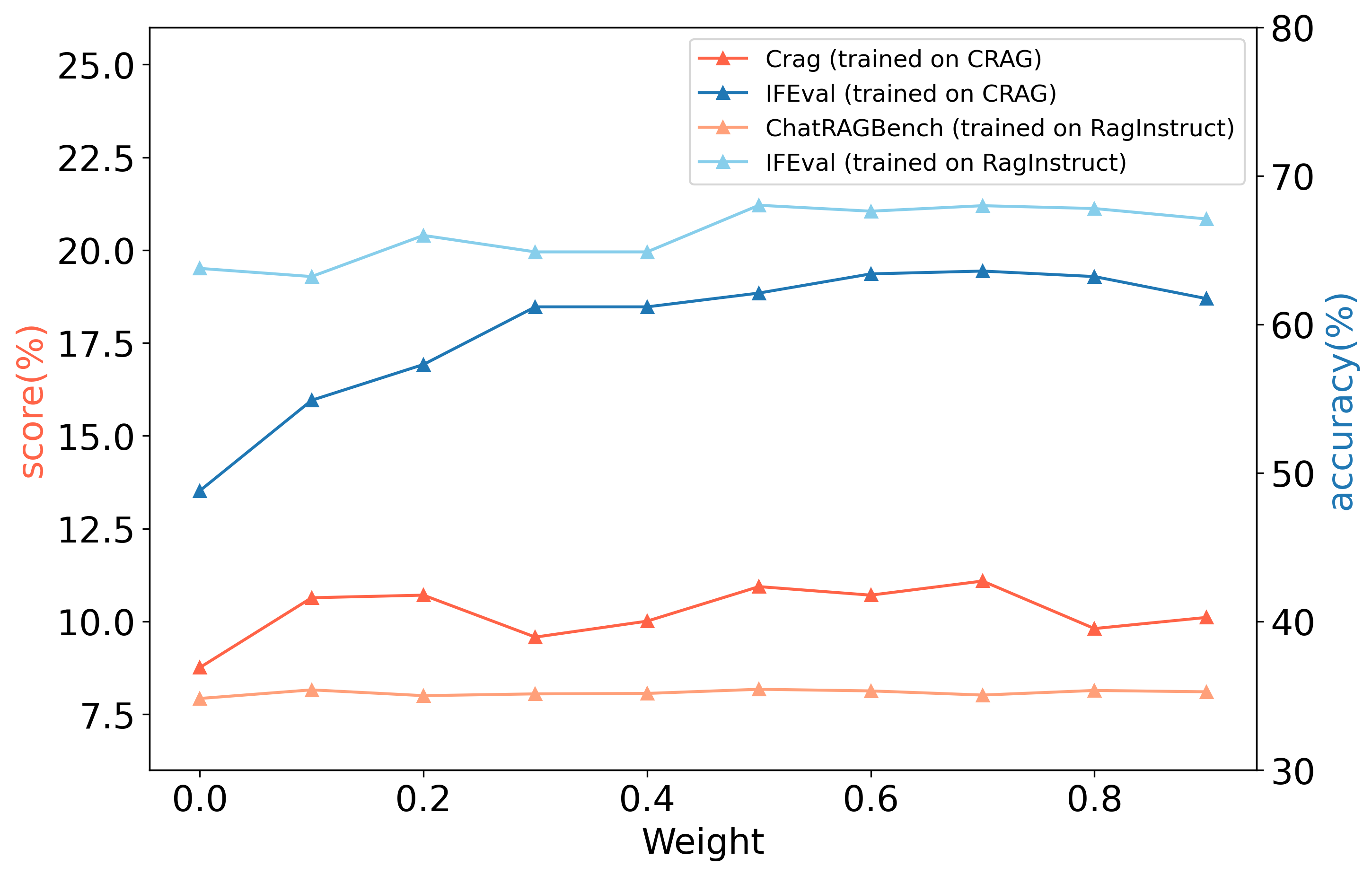}
  \vskip -0.8em
  \caption{Model Performance with Respect to Weight Scaling. Larger loss weights strengthen distribution shift constraints, effectively mitigating forgetting.}
  \label{tab:line_short}
  \vskip -0.8em
\end{figure}

\subsubsection{\textbf{The Impact of Loss Weight.}}
By adjusting the weight parameter $\alpha$ in SelfAug, we can control the strength of the distribution constraints. Higher values of $\alpha$ impose stronger constraints on the model's output distribution, helping to reduce forgetting. However, when $\alpha$ is set too low, the constraint on the model is insufficient, and forgetting is not adequately mitigated; conversely, a high $\alpha$ can hinder downstream task performance. Experimental results show that an $\alpha$ in the range of [0.3, 0.5] typically strikes a good balance between preserving the model’s general abilities and adapting to downstream tasks. As illustrated in Figure \ref{tab:line_short}, increasing $\alpha$ leads to a gradual recovery of the model’s instruction-following ability, demonstrating that SelfAug effectively reduces the divergence between the model's current and original distributions, thereby mitigating catastrophic forgetting. This shows that our approach successfully addresses the root cause of forgetting by keeping the model’s output distribution closer to its initial state while adapting to RAG tasks.

\subsection{Generalizability of SelfAug (RQ4)}

In a RAG scenario, the LLM needs to utilize retrieved documents of varying lengths to answer questions. Therefore, we conducted experiments on model size, LoRA rank, and context length. Additionally, to further validate the effectiveness of our method, we also tested it on tasks with low distribution shift.

\begin{table}[t]
\centering
\caption{Results of Instruction-Following Ability at Different Context Lengths.}
\begin{tabular}{cll}
\toprule
\textbf{Avg Tokens Num} & \textbf{LoRA} & \textbf{SelfAug} \\ 
\midrule
\textbf{2K} tokens & 58.23 & \textbf{63.03} \textcolor{darkgreen}{\scriptsize 4.80$\uparrow$} \\ 
\textbf{4K} tokens & 56.19 & \textbf{62.48} \textcolor{darkgreen}{\scriptsize 6.29$\uparrow$}\\ 
\textbf{6K} tokens & 52.87 & \textbf{55.82} \textcolor{darkgreen}{\scriptsize 2.95$\uparrow$}\\ 
\textbf{8K} tokens & 50.28 & \textbf{57.67} \textcolor{darkgreen}{\scriptsize 7.39$\uparrow$}\\ 
\bottomrule
\end{tabular}
\label{tab:Length}
\vskip -0.6em
\end{table}

\subsubsection{\textbf{Generalizability of SelfAug Across different Context Lengths.}}

As context length increases, model performance on general instruction-following tasks deteriorates due to distribution shift. To investigate this, we examined how training with longer contexts impacts catastrophic forgetting. We progressively expanded the context length by adding more documents and measured instruction-following accuracy at each length, as shown in Table \ref{tab:Length}. As the context length grew from 2K to 8K tokens, instruction-following accuracy dropped from 58.23 to 50.28. Applying SelfAug improved performance, demonstrating its effectiveness in mitigating catastrophic forgetting across all context lengths. When dealing with extremely long input contexts exceeding 32,000 tokens, scalability issues may arise due to the excessive token count, a common challenge in real-world RAG applications. In such cases, attention-based or importance-sampling mechanisms can be used to selectively focus on the most significant tokens in the input, rather than aligning logits across the entire sequence. By identifying and aligning only the most critical tokens, we can significantly reduce computational overhead while maintaining the effectiveness of the alignment strategy.

\subsubsection{\textbf{Generalizability of SelfAug Across different Model Scales.}}
Our investigation into the scalability of SelfAug across different model sizes reveals intriguing patterns, as illustrated in Table \ref{tab:ModelPerformance} through evaluation results on the CRAG benchmark. Contrary to conventional expectations, our experiments demonstrate that the relationship between model size and CRAG performance is not monotonically positive for base models. This counter-intuitive phenomenon can be attributed primarily to the prevalence of hallucination cases in the CRAG dataset, where questions are either inadequately contextualized or fundamentally unanswerable. Particularly noteworthy is our observation that larger base models exhibit diminished performance when encountering such hallucination scenarios, resulting in degraded overall performance metrics.

However, upon fine-tuning with both LoRA and our proposed SelfAug method, we observe a significant paradigm shift in model behavior. The fine-tuned models demonstrate markedly improved capabilities in handling hallucination cases, with performance scaling consistently with model size. Most significantly, our SelfAug approach exhibits superior effectiveness in preserving general capabilities compared to conventional LoRA, effectively mitigating catastrophic forgetting across all model scales. These findings not only validate the scalability of our approach but also underscore its robust performance advantages over existing methods, particularly in addressing the challenging aspects of hallucination management in LLMs.

\begin{table}[t]
    \centering
    \caption{Model Performance Across Different Sizes.}
    \resizebox{0.48\textwidth}{!}{
    \begin{tabular}{r|lll|lll}
    \toprule
    & & \textbf{CRAG} & & & \textbf{IFEval} & \\ 
    \midrule
    \textbf{Size} & \textbf{Base} & \textbf{+LoRA} & \textbf{+SelfAug} & \textbf{Base} & \textbf{+LoRA} & \textbf{+SelfAug} \\
    \midrule
    \textbf{3B}  & -46.82 & 6.37  & \textbf{7.19} \textcolor{darkgreen}{\scriptsize 0.82$\uparrow$} & 61.37 & 49.54  & \textbf{57.86} \textcolor{darkgreen}{\scriptsize 8.32$\uparrow$} \\
    \textbf{7B}  & -13.11 & 8.76  & \textbf{11.24} \textcolor{darkgreen}{\scriptsize 2.48$\uparrow$} & 71.90 & 48.80  & \textbf{62.11} \textcolor{darkgreen}{\scriptsize 13.31$\uparrow$} \\
    \textbf{14B} & -26.29 & 14.31 & \textbf{15.81} \textcolor{darkgreen}{\scriptsize 1.50$\uparrow$} & 79.67 & 45.84 & \textbf{67.47} \textcolor{darkgreen}{\scriptsize 21.63$\uparrow$} \\
    \textbf{32B} & -40.90 & 17.98 & \textbf{19.10} \textcolor{darkgreen}{\scriptsize 1.12$\uparrow$} & 77.45 & 60.81 & \textbf{75.60} \textcolor{darkgreen}{\scriptsize 14.79$\uparrow$} \\
    \textbf{72B} & -20.30 & 19.92 & \textbf{19.93} \textcolor{darkgreen}{\scriptsize 0.01$\uparrow$} & 83.73 & 52.87 & \textbf{62.85} \textcolor{darkgreen}{\scriptsize 9.98$\uparrow$} \\
    \bottomrule
    \end{tabular}}
    \vskip -1em
    \label{tab:ModelPerformance}
\end{table}

\begin{figure}[t]
  \centering
  \includegraphics[width=\linewidth]{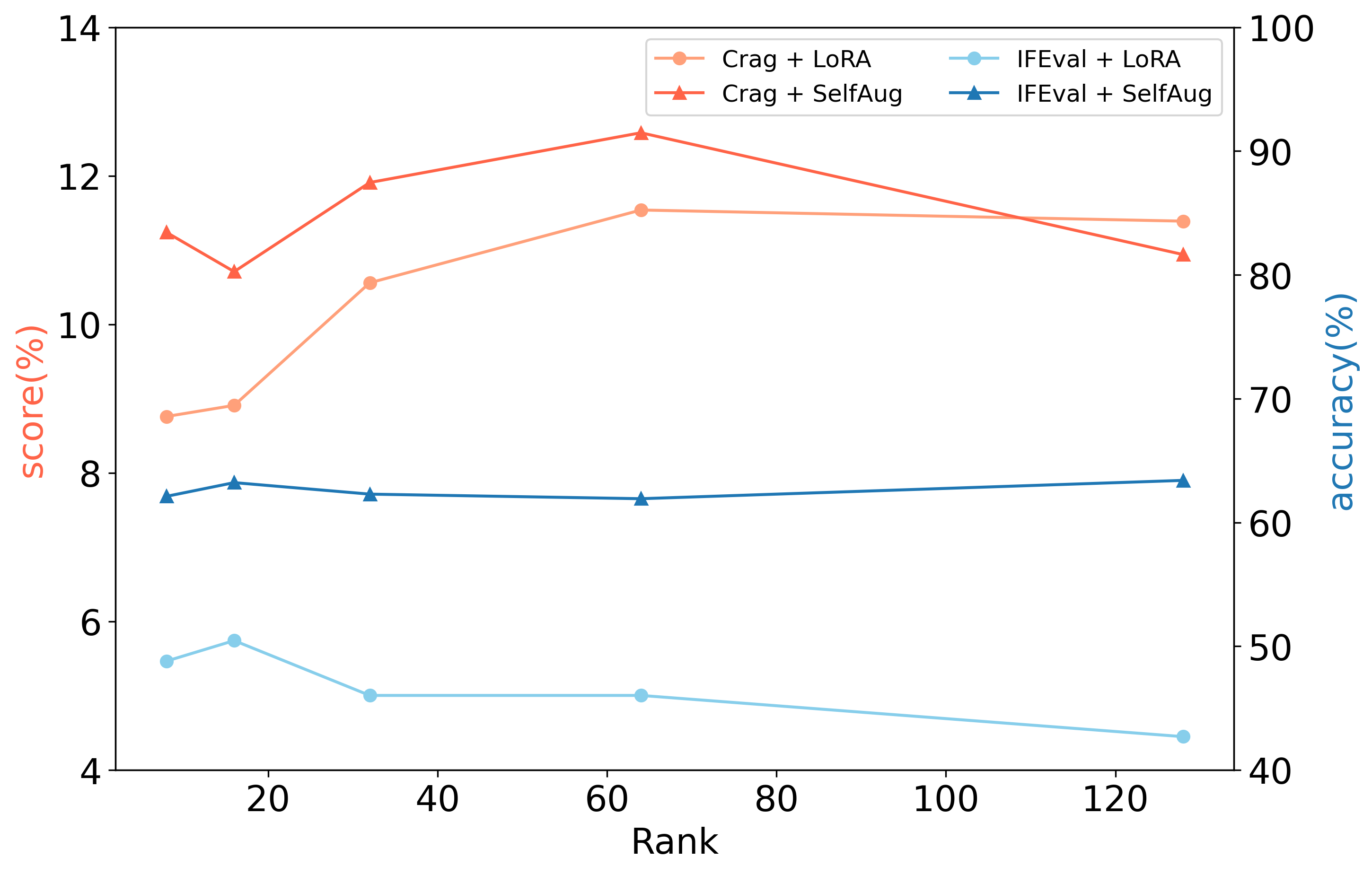}
  \vskip -0.3em
  \caption{Model Performance with Respect to LoRA Rank. Increasing trainable parameters through LoRA rank amplifies catastrophic forgetting severity.}
  \label{tab:rank}
  \vskip -2em
\end{figure}

\begin{table*}[ht]
\centering
\caption{Additional experimental results with Llama-3-8B-Instruct.}
\vskip -0.5em
\resizebox{\textwidth}{!}{
\begin{tabular}{llcccccc}
\toprule
\textbf{Dataset} & \textbf{Method} & \textbf{CRAG} & \textbf{ChatRAGBench} & \textbf{BioASQ} & \textbf{OmniEval} & \textbf{IFEval} & \textbf{AVG} \\
\midrule
- & \textbf{Base} & -8.24 & 32.66 & 52.94 & 45.92 & 75.87 & 50.65 \\
\midrule
& \textbf{LoRA} & 0.00 & 27.62 & 60.00 & 47.42 & 70.98 & 51.20 \\
& \textbf{+MAGPIE} & 0.37 & 32.48 & 57.65 & 47.17 & 71.72 & 51.84 \\
\textbf{CRAG} & \textbf{+SDFT} & 0.54 & 31.99 & 52.94 & 44.61 & 69.50 & 49.86 \\
& \textbf{+Orthogonal} & -1.20 & \textbf{32.57} & \textbf{59.12} & 47.34 & 71.90 & 52.07 \\
& \textbf{+SelfAug (Ours)} & \textbf{0.60} & 32.55 & 58.53 & \textbf{47.42} & \textbf{72.09} & \textbf{52.40} \\
\midrule
& \textbf{LoRA} & -22.47 & 31.91 & 62.35 & 51.78 & 64.33 & 49.83 \\
& \textbf{+MAGPIE} & -11.76 & 33.28 & 60.58 & 50.06 & \textbf{69.50} & 51.51 \\
\textbf{RAG-Instruct} & \textbf{+SDFT} & -19.48 & 29.90 & 56.47 & 43.17 & 66.73 & 47.31 \\
& \textbf{+Orthogonal} & -17.83 & 33.32 & \textbf{62.94} & 49.24 & \textbf{69.50} & 51.22 \\
& \textbf{+SelfAug (Ours)} & \textbf{0.75} & \textbf{36.01} & 62.76 & \textbf{51.84} & \textbf{69.50} & \textbf{54.10} \\
\bottomrule
\end{tabular}}
\label{tab:supplementary_results}
\vskip -1em
\end{table*}

\subsubsection{\textbf{Generalizability of SelfAug Across different Lora Ranks.}}
Having established the correlation between distribution shift and catastrophic forgetting, we investigate the impact of trainable parameters on forgetting severity. Table \ref{tab:main_result} shows that SFT exhibits more severe forgetting than LoRA, suggesting larger trainable parameter sets lead to greater distribution shift. Through controlled experiments with varying LoRA ranks, Figure \ref{tab:rank} reveals that increasing trainable parameters consistently deteriorates instruction-following ability, while our SelfAug method effectively mitigates this across parameter scales. Notably, downstream task performance improves with parameters within an optimal range but degrades beyond a threshold due to redundancy \cite{wang2024lora}. 

\vskip -0.3em
\subsubsection{\textbf{Generalizability of SelfAug On Tasks with Low Distribution Shift.}}

\begin{figure}[t]
  \centering
  \includegraphics[width=\linewidth]{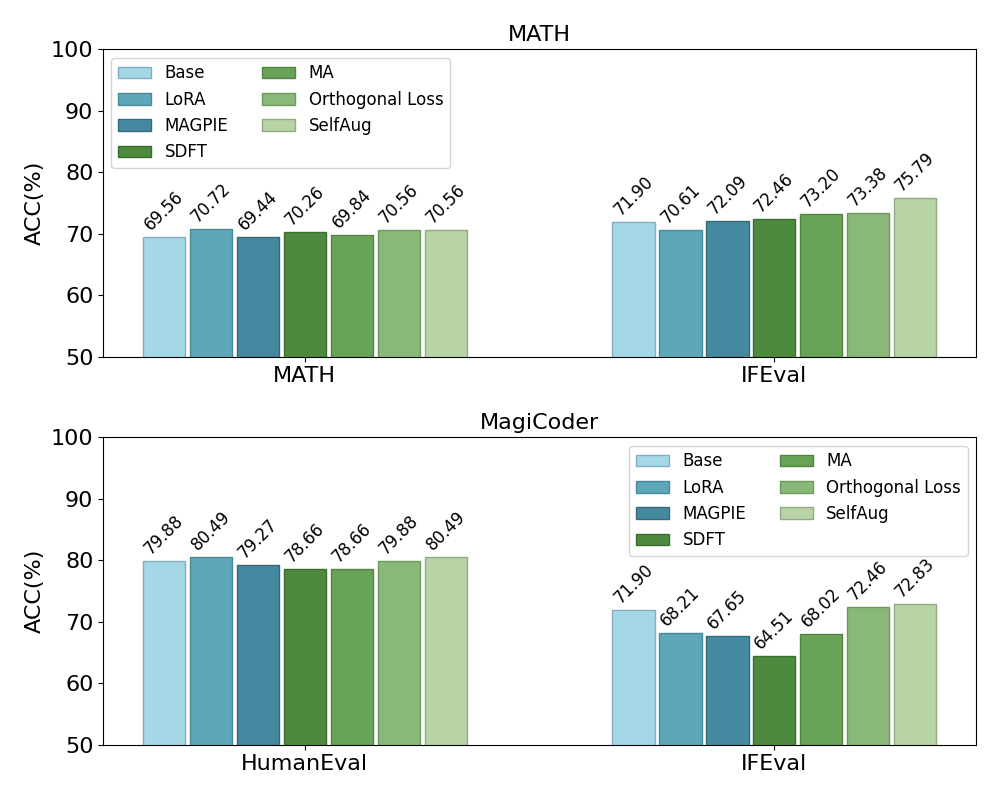}
  \vskip -0.3em
  \caption{Evaluation Results of Math and Code Tasks. SelfAug exhibits forgetting mitigation effectiveness.}
  \label{tab:math}
  \vskip -1em
\end{figure}

To thoroughly assess our approach, we applied Self Aug to mathematical reasoning and code generation tasks, fine-tuning on the MATH and MagiCoder \cite{wei2023magicoder} datasets. As shown in Figure \ref{tab:math}, given the model's extensive pre-training and strong baseline in these areas, additional fine-tuning minimally improved performance, with gains mostly under 1 percentage point. While the conventional LoRA approach showed some decline in instruction-following, SelfAug prevented this and slightly enhanced overall capabilities. This demonstrates SelfAug's effectiveness in maintaining model stability and expanding its benefits across various application domains, even in low distribution shift scenarios.

\subsubsection{\textbf{Additional Experiments with Supplementary Baseline Models.}}

In order to explore the influence of different model architectures on catastrophic forgetting, we conducted additional experiments using Llama-3-8B-Instruct as a supplementary baseline model. This was done to assess the generalizability of our method across a broader range of baseline models. The experimental results, summarized in Table \ref{tab:supplementary_results}, demonstrate that SelfAug consistently mitigates catastrophic forgetting across various architectures. The results highlight that SelfAug consistently outperforms alternative methods, such as LoRA, across various datasets. Notably, the improvements are particularly pronounced on the RAG-Instruct dataset, further validating the robustness of our method in mitigating catastrophic forgetting. These findings indicate that SelfAug is effective not only in preserving general capabilities but also in adapting to different model architectures and experimental setups.

\section{Conclusion}
Our research explores the problem of catastrophic forgetting when fine-tuning language models for retrieval-augmented generation tasks. We find that distribution shift during fine-tuning weakens the model’s general performance, especially its ability to follow instructions. To address this, we propose SelfAug, a method that does not use data replay or change the model architecture, and can be applied to any fine-tuning setting. SelfAug uses only the original training data and aligns the model’s input distributions by constraining input sequence logits. This simple approach reduces distribution shift and helps prevent catastrophic forgetting. Our experiments show that there is a clear link between distribution shift and catastrophic forgetting. SelfAug reduces this shift and preserves model abilities, while matching or exceeding the downstream task performance of standard fine-tuning methods.

\section*{Limitations}

While SelfAug is designed as a plug-and-play approach that integrates seamlessly with both LoRA and full-parameter fine-tuning, we did not conduct extensive experiments on full-parameter settings due to computational constraints. For extremely long input contexts exceeding 32,000 tokens, our method may face scalability issues, as aligning logits across the entire sequence can be costly. We recommend using attention-based or importance-sampling mechanisms to focus only on the most critical tokens. Future work will explore the effectiveness and scalability of SelfAug in full-parameter fine-tuning settings, potentially revealing additional insights into its broader applicability across different training paradigms.

\bibliography{ref}

\appendix

\section{Experimental Setup}
\label{sec:appendix_setup}
\subsection{\textbf{Datasets Details.}} 

\paragraph{RAG Ability Evaluation.} 

The CRAG dataset contains 2.7k question-answer pairs with retrieved reference documents, structured into validation and public test sets. We use the official GPT-4o evaluation protocol, which has been validated against human assessments to minimize bias and ensure the reliability of the evaluation results. The evaluation protocol in CRAG implements a ternary scoring mechanism, where responses are evaluated by GPT-4o to assign scores of 1, -1, and 0 to accurate, incorrect, and missing answers, respectively. The overall score is calculated as the mean score across all responses, with a range of [-1, 1]. 
RAG-Instruct provides a publicly available 40K instruction dataset covering various RAG scenarios. For evaluating multi-turn conversational QA with extensive document contexts, we employ QuAC \cite{choi2018quac}, QReCC \cite{anantha2020open}, and INSCIT \cite{wu2023inscit} following the experimental settings in ChatRAGBench.

\paragraph{Domain-specific RAG Evaluation.} 
BioASQ is a series of international competitions designed to advance large-scale biomedical semantic indexing and question answering. For evaluation, we use Task b from BioASQ 2024 and employ ideal answers as ground truth. OmniEval serves as a RAG benchmark encompassing 5 task categories and 16 financial topics. We rely on GPT-4o for assessment.

\subsection{Implementation Details.} 
The model is trained for 1 epoch with a batch size of 16 and a learning rate of 5e-4. Regarding the RAG-Instruct dataset, we configure the training with a batch size of 512 and a learning rate of 5e-5 over 3 epochs. To mitigate potential model collapse during full parameter fine-tuning at high learning rates, we adopt reduced learning rates of 1e-5 and 5e-6 for CRAG and RAG-Instruct, respectively. Throughout the training process, we employ the AdamW optimizer with a cosine learning rate schedule, setting the weight decay to 0.1 and the warmup ratio to 5\%. In the implementation of MAGPIE, we maintain a mixing ratio of 1:9 between MAGPIE data and original training data. 

\section{Computational Cost Analysis}
SelfAug requires one additional forward pass through the reference model per input during training, which is comparable in cost to SDFT’s answer-rewriting strategy. MAGPIE, by contrast, incurs higher training cost due to the use of extra data for replay. Orthogonal loss introduces an additional loss term, resulting in minimal computational overhead. Overall, the computational cost of SelfAug remains comparable to other strong baselines.

\end{document}